\documentclass[a4paper, 10pt, conference]{ieeeconf}       

\IEEEoverridecommandlockouts                              

\usepackage{graphics} 
\usepackage{epsfig} 
\usepackage{mathptmx} 
\usepackage{times} 
\usepackage{amsmath} 
\usepackage{amssymb}  

\usepackage{geometry}
\makeatletter
\let\MYcaption\@makecaption
\makeatother

\usepackage[font=footnotesize]{subcaption}

\makeatletter
\let\@makecaption\MYcaption
\makeatother

\usepackage{cite}
\usepackage{multirow}

\title{\LARGE \bf Investigating the Effect of LED Signals and Emotional Displays in Human-Robot Shared Workspaces}

\author{Maria Ibrahim$^{1}$, Alap Kshirsagar$^{2}$, Dorothea Koert$^{2,4}$ and Jan Peters$^{2,3,4,5}$
\thanks{\textsuperscript{1} German International University, Egypt.}
\thanks{\textsuperscript{2}Intelligent Autonomous Systems Lab, Department of Computer Science, TU Darmstadt, Germany. {\tt\small alap@robot-learning.de}}
\thanks{\textsuperscript{3}German Research Center for AI (DFKI)}
\thanks{\textsuperscript{4}Centre for Cognitive Science, TU Darmstadt}
\thanks{\textsuperscript{5}Hessian Center for Artificial Intelligence (Hessian.AI), Darmstadt}
\thanks{We thank Hessisches Ministerium für Wissenschaft \& Kunst for the DFKI grant and ``The Adaptive Mind'' grant.}
}

\begin{document}

\maketitle
\thispagestyle{empty}
\pagestyle{empty}

\begin{abstract}

Effective communication is essential for safety and efficiency in human-robot collaboration, particularly in shared workspaces. This paper investigates the impact of nonverbal
communication on human-robot interaction (HRI) by integrating reactive light signals and emotional displays into a robotic system. We equipped a Franka Emika Panda robot with an LED strip on its end effector and an animated facial display on a tablet to convey movement intent through colour-coded signals and facial expressions. We conducted a human-robot collaboration experiment with 18 participants, evaluating three conditions: LED signals alone, LED signals with reactive emotional displays, and LED signals with pre-emptive emotional displays. We collected data through questionnaires and position tracking to assess anticipation of potential collisions, perceived clarity of communication, and task performance. The results indicate that while emotional displays increased the perceived interactivity of the robot, they did not significantly improve collision anticipation, communication clarity, or task efficiency compared to LED signals alone. These findings suggest that while emotional cues can enhance user
engagement, their impact on task performance in shared workspaces is limited.

\end{abstract}

\section{INTRODUCTION}
Robots have transformed industries by surpassing humans in precision, accuracy, durability, and speed~\cite{c2}, making them ideal for tasks that are dangerous, repetitive, or physically demanding~\cite{c3}. 
However, robots still lag behind humans in dexterity and complex decision-making. To address these limitations and optimise performance, collaboration between humans and robots in shared workspaces has become increasingly important. In such environments, effective communication is crucial to ensure both safety and efficiency. This paper explores how nonverbal communication, specifically through reactive light signals and emotional displays, can enhance human-robot interaction (HRI) by enabling more intuitive and responsive behaviour from robotic systems. 

Traditionally, industrial robots have been confined to enclosed areas for safety reasons. Although their movements are predictable, this physical separation limits real-time cooperation. Collaborative robots, or cobots, were designed to overcome this barrier by safely operating alongside humans in shared spaces~\cite{c1}. By combining human adaptability with the precision and resilience of cobots—and enhancing communication through nonverbal cues—these collaborative systems can significantly improve productivity and workflow efficiency.

\begin{figure}[ht]
    \centering
    \includegraphics[width=0.9\linewidth]{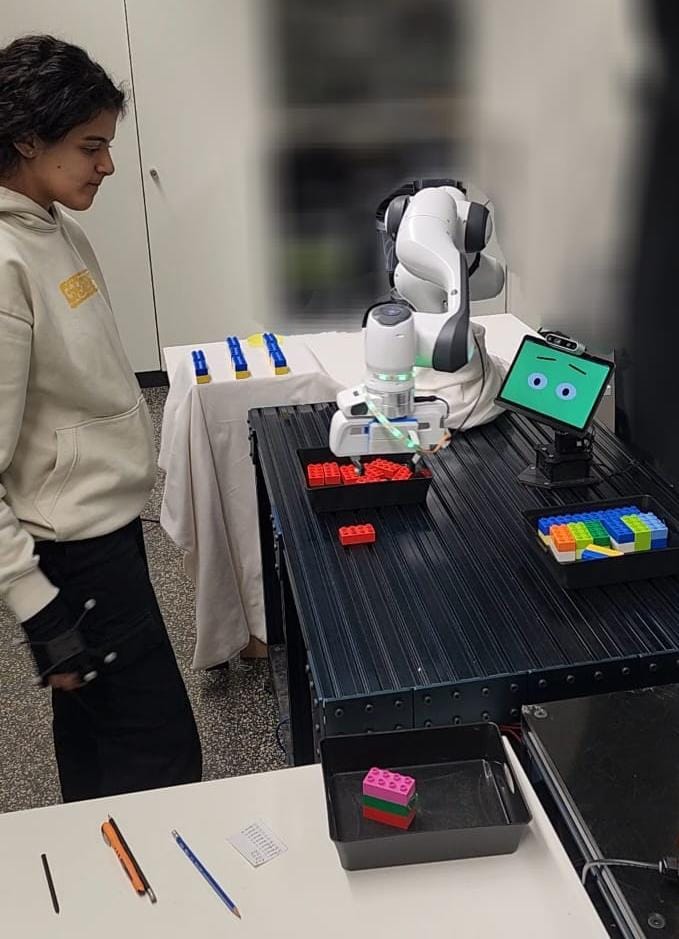}
    \caption{Our experiment setup to study whether emotional display (displayed on a tablet) with LED signals (located on the end-effector) improves HRI in shared workspaces.}
    \label{fig:enter-label}
\end{figure}

Collaborative robots (cobots) are increasingly integrated into industrial settings to enhance productivity and flexibility~\cite{c5}. However, ensuring safety in shared human-robot workspaces remains a critical challenge. To prevent collisions and support seamless collaboration, the robot must convey its intentions through clear and intuitive signals. In many industrial environments, auditory communication can be unreliable due to high levels of background noise. As a result, nonverbal communication becomes not only preferable but essential. By employing visual signals, such as lights, gestures, or expressive displays, robots can communicate their actions and intentions, potentially improving safety and coordination in shared workspaces. This paper focuses on enhancing HRI in a shared workspace by testing the effectiveness of a light-based signal using an LED strip around the end effector combined with an emotional display to show the robot's collision avoidance intent.

As described in Section~\ref{sec:related-work}, there is a significant deficit in research on the use of emotional displays for robots in shared workspaces between humans and robots. Although previous studies have explored nonverbal cues, such as gaze behaviour and light-based signals, to improve communication in shared workspaces, the potential of emotional displays to enhance human-robot collaboration remains largely unexplored. Emotional displays could provide a more intuitive and natural way for robots to communicate their intentions, improving human comfort and efficiency in collaborative environments.

The experiment design, described in Section~\ref{sec:method}, involves human participants working alongside a robot to perform a block assembly task. The setup mimics the scenarios of a shared workspace in the real world. To assess the effectiveness of light-based signals and emotional displays to convey the robot's collision avoidance intent, we evaluate three distinct conditions. The first condition involves only LED signals, the second condition combines LED signals with reactive emotional displays, and the third condition incorporates preemptive emotional displays alongside LED signals. The results (described in Section~\ref{sec:results}) show that emotional displays significantly increase perceived robot interactivity compared to LED signals alone ($p=0.019$), but do not affect collision anticipation, clarity, or efficiency. Our findings have implications for designing multimodal cues and adaptive intent prediction for intuitive cobots, to bridge the gap between engagement and performance in shared workspaces.

\section{RELATED WORK}
\label{sec:related-work}

Light-based signals have emerged as a robust method for conveying robot intent in collaborative environments. Lemasurier et al.~\cite{Lema} investigated visual cues, including LED bracelets, gaze, arm light, head pan, forearm movement, and gripper movement. Their findings highlight the effectiveness of LED bracelets in signalling motion intent, demonstrating high noticeability and minimal confusion for human collaborators compared to other signals. Similarly, Cha et al.~\cite{Cha2018} categorise visual signalling methods, such as blinker lights and ambient light signals, as effective in conveying a robot's navigational intent, with applications in autonomous vehicles and hallway navigation \cite{Cha2018, Szafir2015}. However, these signals are often designed for independent robot movement rather than collaborative tasks, and they may lack the emotional context needed to enhance user engagement, prompting the exploration of complementary modalities.

Beyond LED-based approaches, light projection and augmented reality (AR) have been explored for intent communication. Chadalavada et al.~\cite{Chadalavada2015} demonstrated that projecting a robot's intended path onto the environment improves human anticipation of robot movements in shared spaces. However, projection-based methods are limited by their reliance on flat surfaces and susceptibility to occlusion, making them less effective for complex 3D movements, such as those of a robotic arm \cite{Andersen2016}. As explored by Walker et al. \cite{Walker2018}, AR-based signalling uses virtual cues such as arrows or gaze to guide human attention, improving task coordination. Despite their potential, AR systems require users to wear specialised equipment, which may not be practical in dynamic workspaces and poses safety risks if the equipment is unavailable \cite{Walker2018}. These limitations underscore the need for scalable and intuitive visual signalling methods, such as our study's proposed LED and emotional display system.

Emotional displays offer a promising avenue for making HRI more intuitive. Yang et al.~\cite{Yang} explored nonverbal emotional expressions (e.g. happy or angry faces) in a tic-tac-toe game with humans, reporting a 20\% increase in task efficiency and a 30\% reduction in errors when emotional cues were used. While their work highlights the potential of emotional signals to improve collaboration, it focuses on interactive tasks with the robot rather than shared workspaces. Our study addresses this gap by integrating emotional displays with a collaborative robotic arm in a dynamic, collision-prone environment. Using a tablet-based facial interface, our approach mimics human-like expressiveness, aiming to enhance safety and engagement. Furthermore, studies like those by Breazeal~\cite{Breazeal2003} emphasise that emotional displays can foster trust and predictability in HRI. Still their application in industrial settings remains underexplored, motivating our focus on collision avoidance in shared workspaces.

Motion-based signals, which involve physical movements to convey intent, have also been studied for their robustness in varying environmental conditions. Dragan et al.~\cite{Dragan2013} developed cost functions to optimise robot motion for legibility, ensuring movements are intuitive and aligned with human expectations. Anticipatory motions, such as slight arm adjustments before a task, improve human understanding of the robot's intent~\cite{Dragan2013}. Furthermore, Szafir et al.~\cite{Szafir2014} explored whole-body movements and gestures in assistive robots, establishing guidelines for expressive motion across platforms. Unlike visual signals, motion-based cues do not rely on external lighting or equipment, making them suitable for diverse settings. However, their effectiveness in close-proximity collaboration remains less studied, and they risk being perceived as threatening if not carefully designed \cite{Szafir2014}. Our study complements these findings by focusing on visual and emotional signals, which may be less intrusive in shared workspaces.

Broader frameworks for intent communication provide additional context for our work. Pascher et al.~\cite{Pasch} proposed a model that classifies robot intent into motion, attention, state, and instruction types, highlighting the role of spatial and temporal clarity in HRI. Their framework informs our dual-modality approach, which combines LED signals for motion intent with emotional displays to communicate the robot's reaction to the human. This integration aims to enhance both transparency and participation in human-robot collaboration. Other nonverbal cues, such as gaze behaviours~\cite{Admoni, Moon2014}, have been extensively studied for intent prediction but face practical challenges, including hardware complexity and calibration problems. In contrast, our emotional display and LED system offer a scalable and visually intuitive solution. Building on previous work, this study advances HRI by evaluating a novel combination of light and emotional signals for conveying collision avoidance intent of a collaborative robot. 

\section{METHODOLOGY}
\label{sec:method}

This section describes the experimental framework used to evaluate the efficacy of nonverbal communication in HRI within shared workspaces. We used a within subjects experiment design that involved a collaborative block assembly task with a Franka Emika Panda robot equipped with an LED strip and a tablet-based emotional display. We tested three experimental conditions: LED signals alone, LED signals with reactive emotional displays, and LED signals with pre-emptive emotional displays. We collect data through questionnaires and position tracking to assess collision anticipation, clarity of communication, and performance of the task. 

\subsection{Research Question and Hypothesis}
The primary research question guiding this study is: \emph{How do emotional displays combined with LED signals affect human behaviour and perception in human-robot collaboration compared to LED signals alone?} This question aims to determine whether the integration of emotional displays improves the intuitiveness and safety of HRI in shared workspaces, particularly by conveying the intent of the robot to avoid collisions, ie stopping or continuing when a human hand is detected in the robot's path.

To address this question, the following hypotheses were formulated:

\begin{enumerate}
    \item H1: Emotional displays combined with LED signals will increase participants’ anticipation of potential collisions with the robot compared to LED signals alone.
    \item H2: Emotional displays combined with LED signals will increase the ability of the robot to express its intent clearly to humans when working in a shared workspace compared to LED signals alone.
    \item H3: Emotional displays combined with LED signals will improve human perception compared to LED signals alone.
    \item H4: Emotional displays combined with LED signals will affect human behaviour, as evidenced by changes in collision rates and task execution time, compared to LED signals alone.
\end{enumerate}

\subsection{Experiment Design}
\label{sec:experiment-design}
\begin{figure}[h]
    \centering
    \includegraphics[width=\linewidth]{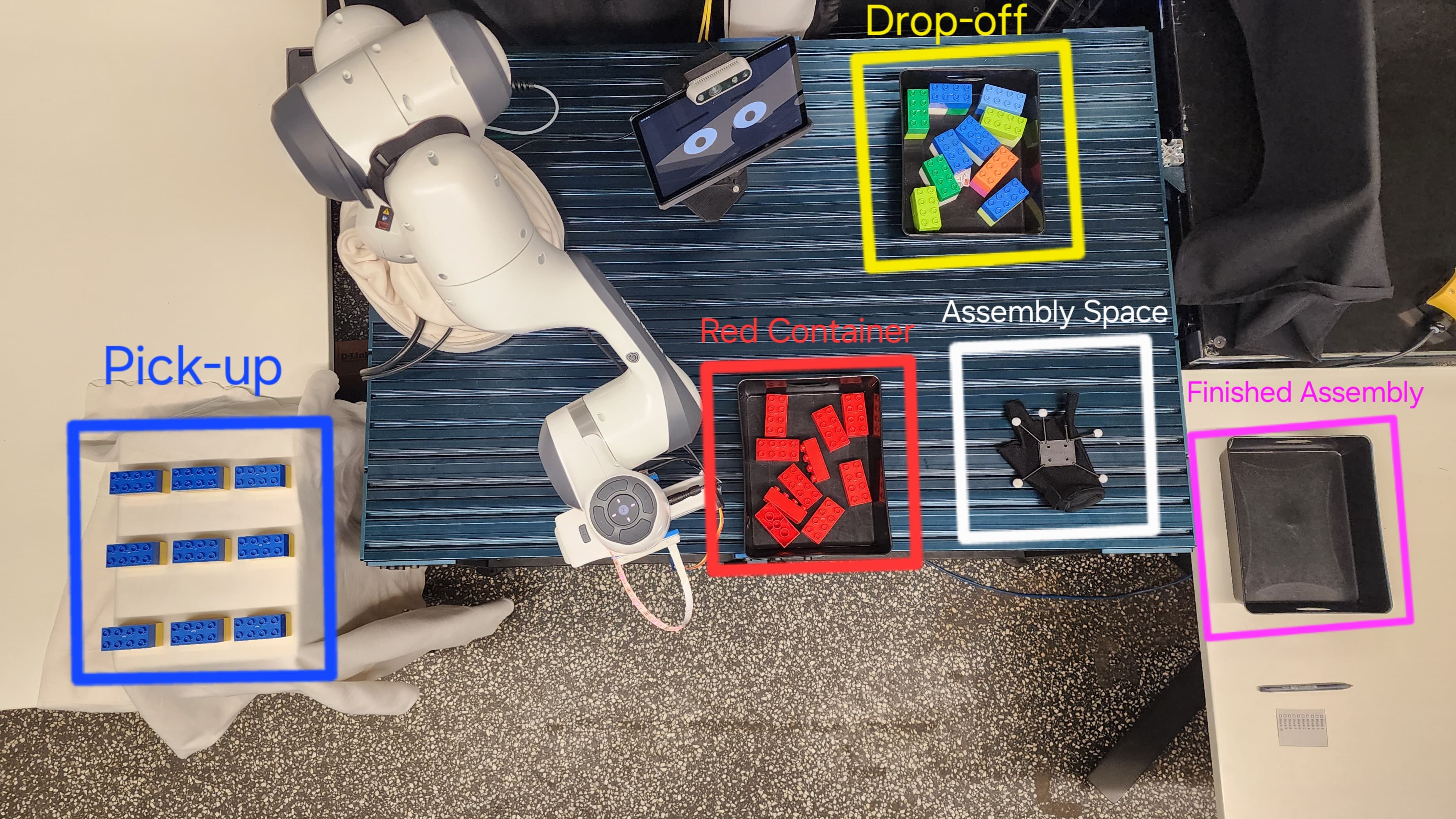}
    \caption{Top view of the experiment setup. The robot picks up blocks from the ``Pick-up" area and puts them in the ``Drop-off" area. The participant has to pick blocks from the ``Red Container" and ``Drop-off" areas, assemble them in the ``Assembly Space" area, place the finished assembly in the ``Finished Assembly" area, tick the box on the sheet next to that area, and write the color of the assembled top block on the sheet.}
    \label{fig:Setup}
\end{figure}

In our experiment, participants performed a collaborative assembly task with the robot. The experiment setup is shown in Figure~\ref{fig:Setup}. The task involved the following steps for the participants: 
\begin{enumerate}
    \item Take a red block from the container and place it in the assembly space.
    \item Take a `double' block from the drop-off container and insert it on top of the red block in the assembly space. 
    \item Place the assembled blocks in the finished assembly container.
    \item Tick the corresponding box on the provided sheet and write the colour of the assembled top block.
    \item Repeat steps 1-4 until no red blocks are left in the container.
\end{enumerate}
The robot transferred individual double blocks from the pickup area and put them in the drop-off container. During movement between the pick-up and drop-off zones, the robot operated in two distinct modes: yielding mode and unyielding mode. In yield mode, the robot stopped if a human hand entered its path and continued to move towards its target only after the hand moved out of its path. In unyielding mode, the robot continued its motion towards its target without stopping, even if a human hand was detected in its path. The mode was set before the start of each movement between the pick-up and drop-off zone. The robot exhibited nonverbal signals to communicate its mode to the human.

\begin{figure}[ht]
    \centering
    \begin{subfigure}[b]{0.45\linewidth}
        \includegraphics[width=\textwidth]{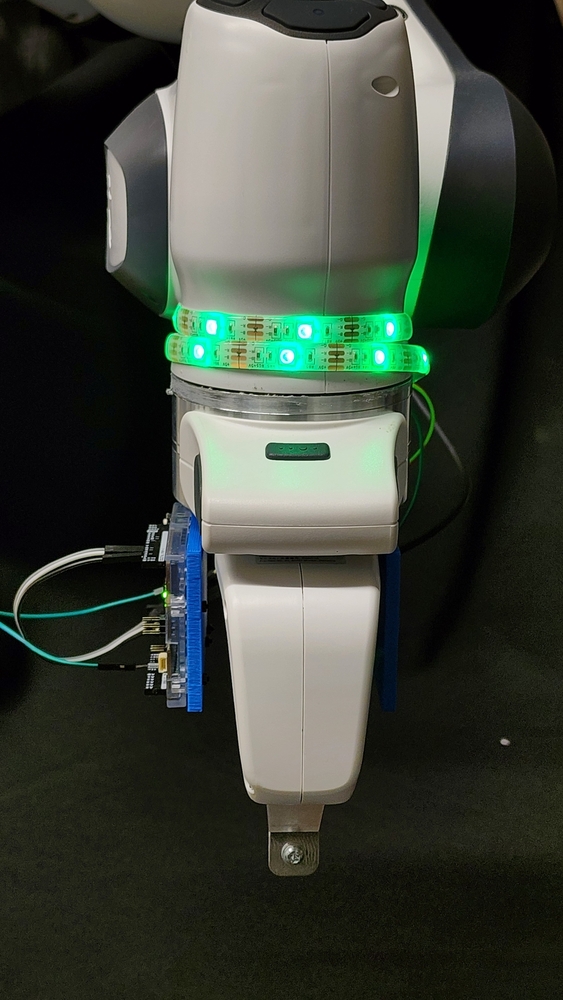}
        \caption{}
        \label{fig:Green_LED}
    \end{subfigure}
    \hfill
    \begin{subfigure}[b]{0.45\linewidth}
        \includegraphics[width=\textwidth]{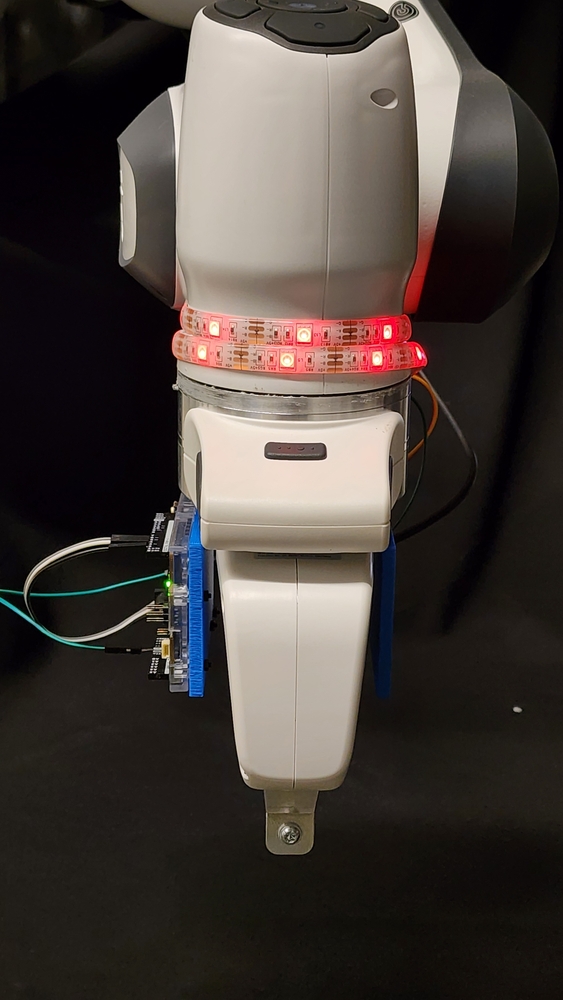}
        \caption{}
        \label{fig:Red_LED}
    \end{subfigure}
    \caption{The two modes of the LED strip used in our experiment. (a) Green LED signal (yielding mode) means that the robot will stop if the participant’s hand is in the robot’s path. (b) Red LED signal (unyielding mode) means that the robot will not stop even if the participant’s hand is in the path}
    \label{fig:LED}
\end{figure}

In our experiment, we evaluated three nonverbal signal conditions. Condition 1 is a baseline with only LED signals. The LED signal has 2 modes: red and green, as shown in Figure~\ref{fig:LED}. The red signal corresponds to the non-yielding mode, i.e. the robot will not stop even if the human’s hand is in the robot's path, and the green signal corresponds to the yielding mode, i.e. the robot will stop if the human’s hand is in the robot's path. The signal changes before the robot starts its movement between the pick-up and drop-off zones and remains constant throughout the movement. 

\begin{figure}[htbp]
    \centering
    \begin{subfigure}[b]{0.3\linewidth}
        \includegraphics[width=\textwidth]{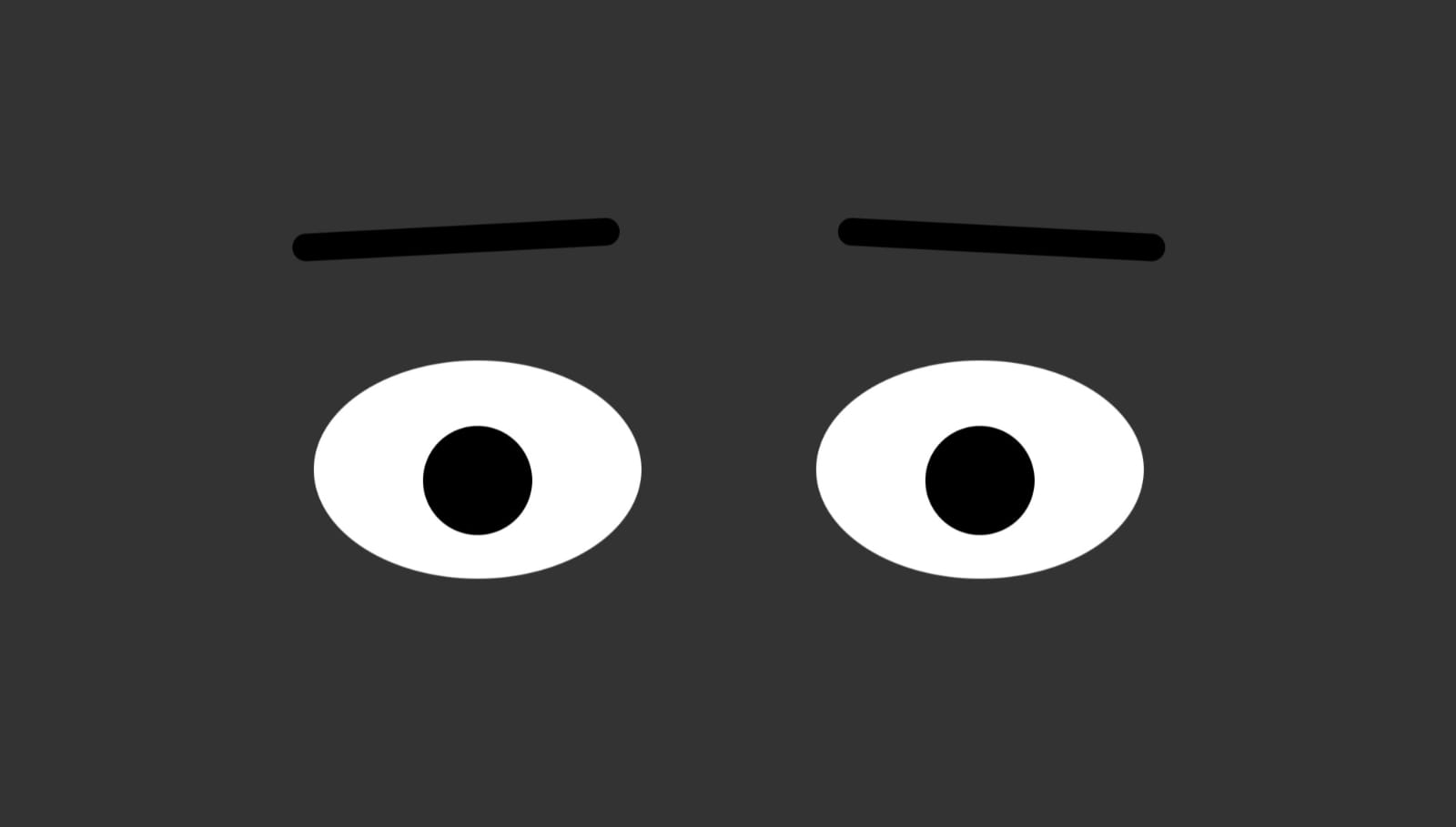}
        \caption{}
        \label{fig:graph1}
    \end{subfigure}
    \hfill
    \begin{subfigure}[b]{0.3\linewidth}
        \includegraphics[width=\textwidth]{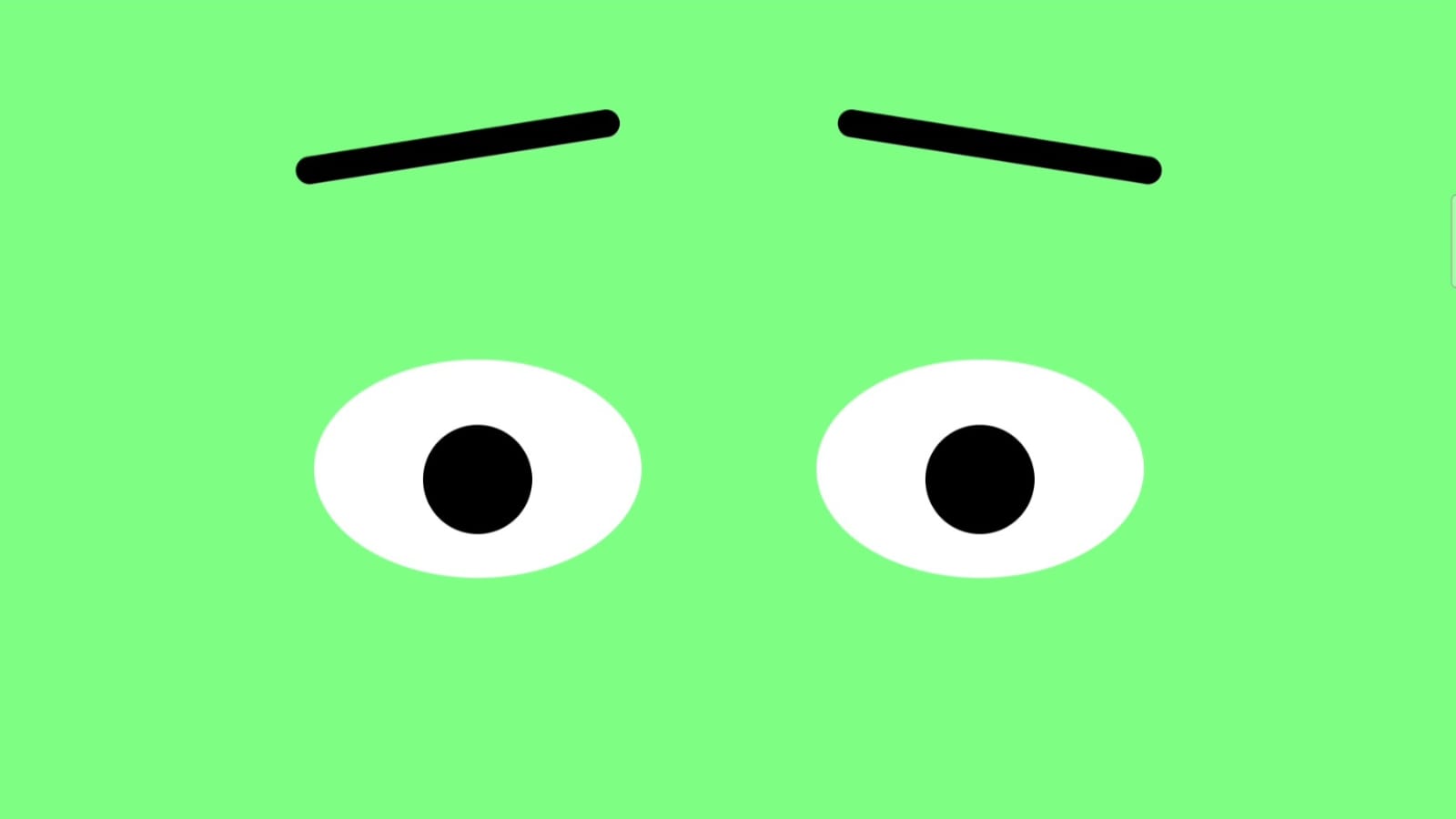}
        \caption{}
        \label{fig:graph2}
    \end{subfigure}
    \hfill
    \begin{subfigure}[b]{0.3\linewidth}
        \includegraphics[width=\textwidth]{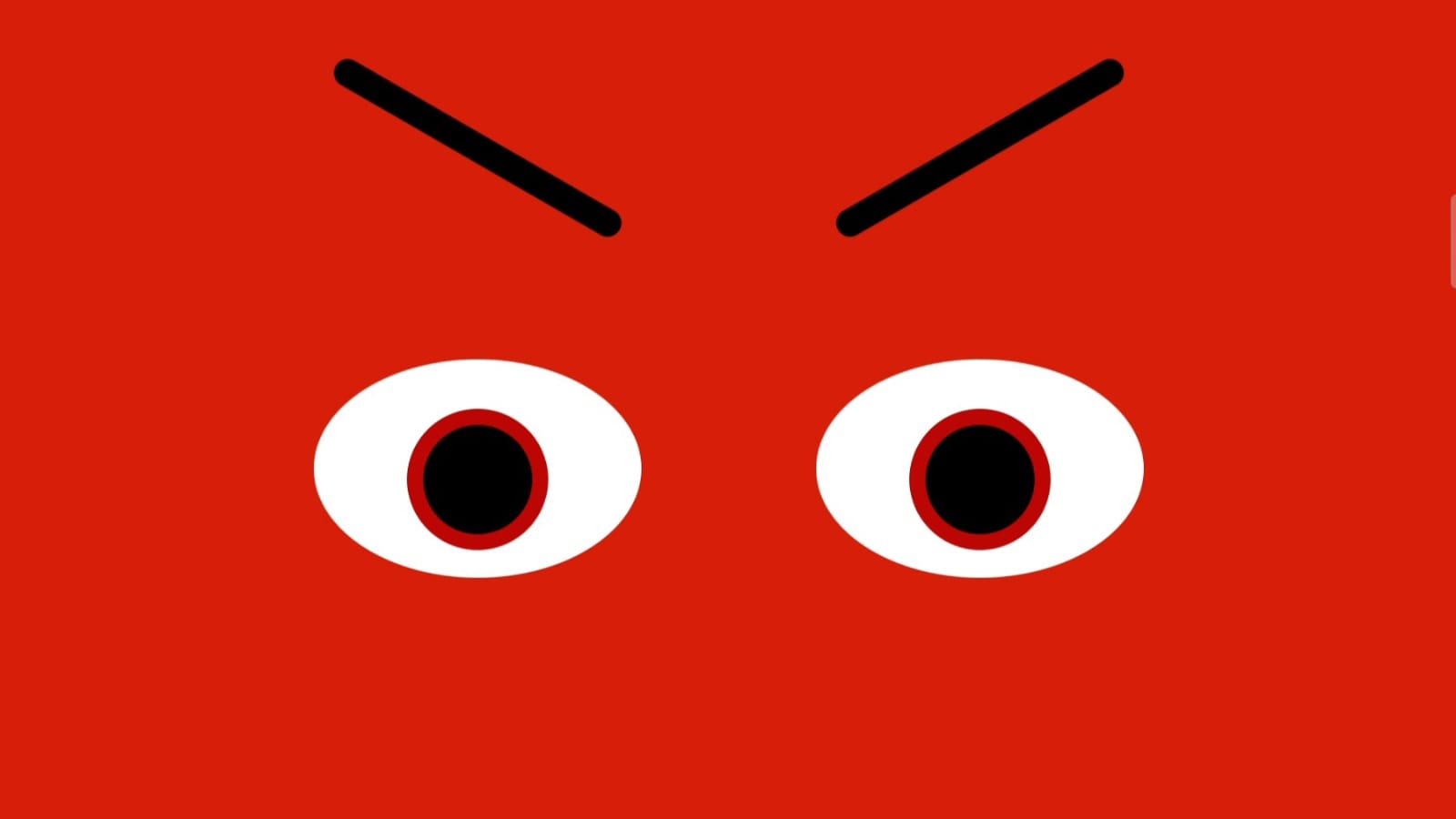}
        \caption{}
        \label{fig:graph3}
    \end{subfigure}
    \caption{The robot's facial expressions used in our experiment. (a) Neutral expression with a grey background. (b) Watchful expression with green background. (c) Angry expression with red background. }
    \label{fig:faces}
\end{figure}

Condition 2 consists of the same LED signals as Condition 1, combined with a reactive emotional display. 
The default state of the robot's face is neutral with a grey background, as shown in Figure~\ref{fig:faces} (a). When there is a potential collision while the LED is green (yielding mode), the face changes to a watchful expression and the background turns green, as shown in Figure~\ref{fig:faces} (b). This indicates that the robot is attentive and that it is safe to move in the robot's path, as the robot will slow down and stop. After a couple of seconds, the face returns to its default state. If there is a potential collision while the LED is red (unyielding mode), the face instead changes to an angry expression and the background turns red, as shown in Figure~\ref{fig:faces} (c). This indicates that the robot is angry that the human hand is in its way and that it will not stop. After a couple of seconds, the background and expression return to their default grey and neutral appearance.

Condition 3 consists of the same LED signals as Condition 1 combined with a pre-emptive emotional display. The display background color changes to match the colour of the LED before the robot begins its movement. The facial expression, with a neutral expression as the default, changes similarly to Condition 2 but without altering the background colour.

The experiment followed a within-subjects design, with each participant experiencing the three conditions. To minimise order effects, the presentation order of the conditions was randomised and counterbalanced. Before the main rounds, the participants performed a practice round to help get familiar with the setup, and this round included LED signals only.

\subsection{Components}

\subsubsection{Robot} We used a Franka-Emika Panda robot, equipped with a joint impedance controller for safe interaction. If the human collided with the robot, the robot would not apply excessive force. However, we did not observe any collisions in our experiment. We used a time-scaled trajectory generator to implement the yielding mode by reducing the robot's speed to zero if the human hand was detected in the robot's path. We added an LED strip on the end effector to signal the robot's collision avoidance intent, i.e. yielding or unyielding mode.

\subsubsection{Emotional Display} We used a Samsung S6 tablet as the robot’s head to display animated facial expressions (neutral, watchful, angry) and background colours (grey, green, red) with a web-based interface. Based on feedback from our pilot trials, we mounted the tablet head on the table rather than at a higher, human-like position. At the higher placement, the head was outside the human's main visual field when they performed the assembly task.

\subsubsection{Tracking System} We used an OptiTrack motion tracking system with six cameras to track participants' hand positions using reflective markers on a glove.

\subsubsection{Integration} We used an Arduino Uno R4 WiFi board to control the LED strip. All components were integrated using the Robot Operating System (ROS).

\subsection{Experiment Procedures}

Upon arrival, participants were first asked to read and sign the consent form if they agreed to participate in the study. Following their consent, they were instructed to read the experiment instructions. The instructions outlined the experimental task. Importantly, the instructions described the functioning of the LED signals. This approach ensured consistency among all participants and minimised the variability that could arise from verbal explanations, thereby controlling for potential confounding variables. The participants were instructed to use only one hand to assemble the blocks and mark the provided sheet so that the OptiTrack system could track the hand's position. They wore gloves with reflective markers on their hand throughout the experiment.  

The participants were instructed to perform the assembly task, described in Section~\ref{sec:experiment-design}, for 10 blocks as quickly as possible in each round. This was stated to simulate a real-world collaborative scenario where efficiency is prioritised. The participants’ focus on completing the task as quickly as possible created a realistic environment to test whether they could notice the robot's emotion display and interpret its signals while working under time pressure.

\subsection{Data Collection}
We collected the participants' self-reported gender, age group, and familiarity with the robot. We used an individual identification code generated by participants to store their data anonymously. After completion of a round, participants were asked to answer a questionnaire, shown in Table~\ref{Tab:Questionnaire} below, about their experience. The responses to the questions were on a Likert scale from 1 to 9, with 1 being ``Strongly disagree'' and 9 being ``Strongly agree''. Once all rounds were completed, participants were asked to answer four final questions: Q1 -- In which round did the behaviour of the robot make you feel the most disturbed?; Q2 -- In which round the robot's intent is most understandable to you?; Q3 -- In which round did you feel most uncomfortable with the robot?; Q4 -- Do you have anything else to say?

\begin{table}[h]
\centering
\caption{Questionnaire given to the participants after every round}
\begin{tabular}{|ccccccccccc|}
\hline
\multicolumn{11}{|c|}{Question 1: ANTICIPATION OF DANGER} \\ \hline
\multicolumn{11}{|c|}{“I was able to anticipate potential collisions with the robot”} \\ \hline
\multicolumn{1}{|c|}{\multirow{2}{*}{\begin{tabular}[c]{@{}c@{}}Strongly\\ disagree\end{tabular}}} &
  \multicolumn{1}{c|}{\multirow{2}{*}{1}} &
  \multicolumn{1}{c|}{\multirow{2}{*}{2}} &
  \multicolumn{1}{c|}{\multirow{2}{*}{3}} &
  \multicolumn{1}{c|}{\multirow{2}{*}{4}} &
  \multicolumn{1}{c|}{\multirow{2}{*}{5}} &
  \multicolumn{1}{c|}{\multirow{2}{*}{6}} &
  \multicolumn{1}{c|}{\multirow{2}{*}{7}} &
  \multicolumn{1}{c|}{\multirow{2}{*}{8}} &
  \multicolumn{1}{c|}{\multirow{2}{*}{9}} &
  \multirow{2}{*}{\begin{tabular}[c]{@{}c@{}}Strongly\\ agree\end{tabular}} \\
\multicolumn{1}{|c|}{} &
  \multicolumn{1}{c|}{} &
  \multicolumn{1}{c|}{} &
  \multicolumn{1}{c|}{} &
  \multicolumn{1}{c|}{} &
  \multicolumn{1}{c|}{} &
  \multicolumn{1}{c|}{} &
  \multicolumn{1}{c|}{} &
  \multicolumn{1}{c|}{} &
  \multicolumn{1}{c|}{} &
   \\ \hline
\multicolumn{11}{|c|}{Question 2: INTENT COMMUNICATION} \\ \hline
\multicolumn{11}{|c|}{“The robot communicated its intent clearly”} \\ \hline
\multicolumn{1}{|c|}{\multirow{2}{*}{\begin{tabular}[c]{@{}c@{}}Strongly\\ disagree\end{tabular}}} &
  \multicolumn{1}{c|}{\multirow{2}{*}{1}} &
  \multicolumn{1}{c|}{\multirow{2}{*}{2}} &
  \multicolumn{1}{c|}{\multirow{2}{*}{3}} &
  \multicolumn{1}{c|}{\multirow{2}{*}{4}} &
  \multicolumn{1}{c|}{\multirow{2}{*}{5}} &
  \multicolumn{1}{c|}{\multirow{2}{*}{6}} &
  \multicolumn{1}{c|}{\multirow{2}{*}{7}} &
  \multicolumn{1}{c|}{\multirow{2}{*}{8}} &
  \multicolumn{1}{c|}{\multirow{2}{*}{9}} &
  \multirow{2}{*}{\begin{tabular}[c]{@{}c@{}}Strongly\\ agree\end{tabular}} \\
\multicolumn{1}{|c|}{} &
  \multicolumn{1}{c|}{} &
  \multicolumn{1}{c|}{} &
  \multicolumn{1}{c|}{} &
  \multicolumn{1}{c|}{} &
  \multicolumn{1}{c|}{} &
  \multicolumn{1}{c|}{} &
  \multicolumn{1}{c|}{} &
  \multicolumn{1}{c|}{} &
  \multicolumn{1}{c|}{} &
   \\ \hline
\multicolumn{11}{|c|}{Question 3: PERCEIVED COMPETENCE} \\ \hline
\multicolumn{11}{|c|}{\begin{tabular}[c]{@{}c@{}}Using the scales provided, how closely are the\\ words associated with your impression of the robot?\end{tabular}} \\ \hline
\multicolumn{1}{|c|}{\multirow{2}{*}{\begin{tabular}[c]{@{}c@{}}Strongly\\ disagree\end{tabular}}} &
  \multicolumn{9}{c|}{Knowledgeable} &
  \multirow{2}{*}{\begin{tabular}[c]{@{}c@{}}Strongly\\ agree\end{tabular}} \\ \cline{2-10}
\multicolumn{1}{|c|}{} &
  \multicolumn{1}{c|}{1} &
  \multicolumn{1}{c|}{2} &
  \multicolumn{1}{c|}{3} &
  \multicolumn{1}{c|}{4} &
  \multicolumn{1}{c|}{5} &
  \multicolumn{1}{c|}{6} &
  \multicolumn{1}{c|}{7} &
  \multicolumn{1}{c|}{8} &
  \multicolumn{1}{c|}{9} &
   \\ \hline
\multicolumn{1}{|c|}{\multirow{2}{*}{\begin{tabular}[c]{@{}c@{}}Strongly\\ disagree\end{tabular}}} &
  \multicolumn{9}{c|}{Interactive} &
  \multirow{2}{*}{\begin{tabular}[c]{@{}c@{}}Strongly\\ agree\end{tabular}} \\ \cline{2-10}
\multicolumn{1}{|c|}{} &
  \multicolumn{1}{c|}{1} &
  \multicolumn{1}{c|}{2} &
  \multicolumn{1}{c|}{3} &
  \multicolumn{1}{c|}{4} &
  \multicolumn{1}{c|}{5} &
  \multicolumn{1}{c|}{6} &
  \multicolumn{1}{c|}{7} &
  \multicolumn{1}{c|}{8} &
  \multicolumn{1}{c|}{9} &
   \\ \hline
\multicolumn{1}{|c|}{\multirow{2}{*}{\begin{tabular}[c]{@{}c@{}}Strongly\\ disagree\end{tabular}}} &
  \multicolumn{9}{c|}{Responsive} &
  \multirow{2}{*}{\begin{tabular}[c]{@{}c@{}}Strongly\\ agree\end{tabular}} \\ \cline{2-10}
\multicolumn{1}{|c|}{} &
  \multicolumn{1}{c|}{1} &
  \multicolumn{1}{c|}{2} &
  \multicolumn{1}{c|}{3} &
  \multicolumn{1}{c|}{4} &
  \multicolumn{1}{c|}{5} &
  \multicolumn{1}{c|}{6} &
  \multicolumn{1}{c|}{7} &
  \multicolumn{1}{c|}{8} &
  \multicolumn{1}{c|}{9} &
   \\ \hline
\multicolumn{1}{|c|}{\multirow{2}{*}{\begin{tabular}[c]{@{}c@{}}Strongly\\ disagree\end{tabular}}} &
  \multicolumn{9}{c|}{Capable} &
  \multirow{2}{*}{\begin{tabular}[c]{@{}c@{}}Strongly\\ agree\end{tabular}} \\ \cline{2-10}
\multicolumn{1}{|c|}{} &
  \multicolumn{1}{c|}{1} &
  \multicolumn{1}{c|}{2} &
  \multicolumn{1}{c|}{3} &
  \multicolumn{1}{c|}{4} &
  \multicolumn{1}{c|}{5} &
  \multicolumn{1}{c|}{6} &
  \multicolumn{1}{c|}{7} &
  \multicolumn{1}{c|}{8} &
  \multicolumn{1}{c|}{9} &
   \\ \hline
\multicolumn{1}{|c|}{\multirow{2}{*}{\begin{tabular}[c]{@{}c@{}}Strongly\\ disagree\end{tabular}}} &
  \multicolumn{9}{c|}{Competent} &
  \multirow{2}{*}{\begin{tabular}[c]{@{}c@{}}Strongly\\ agree\end{tabular}} \\ \cline{2-10}
\multicolumn{1}{|c|}{} &
  \multicolumn{1}{c|}{1} &
  \multicolumn{1}{c|}{2} &
  \multicolumn{1}{c|}{3} &
  \multicolumn{1}{c|}{4} &
  \multicolumn{1}{c|}{5} &
  \multicolumn{1}{c|}{6} &
  \multicolumn{1}{c|}{7} &
  \multicolumn{1}{c|}{8} &
  \multicolumn{1}{c|}{9} &
   \\ \hline
\multicolumn{1}{|c|}{\multirow{2}{*}{\begin{tabular}[c]{@{}c@{}}Strongly\\ disagree\end{tabular}}} &
  \multicolumn{9}{c|}{Reliable} &
  \multirow{2}{*}{\begin{tabular}[c]{@{}c@{}}Strongly\\ agree\end{tabular}} \\ \cline{2-10}
\multicolumn{1}{|c|}{} &
  \multicolumn{1}{c|}{1} &
  \multicolumn{1}{c|}{2} &
  \multicolumn{1}{c|}{3} &
  \multicolumn{1}{c|}{4} &
  \multicolumn{1}{c|}{5} &
  \multicolumn{1}{c|}{6} &
  \multicolumn{1}{c|}{7} &
  \multicolumn{1}{c|}{8} &
  \multicolumn{1}{c|}{9} &
   \\ \hline
\end{tabular}
\label{Tab:Questionnaire}
\end{table}

\subsection{Participants}
A total of 27 participants participated in the study. Data from 9 of them had to be excluded due to software malfunctions, which led to issues such as the LED or the tablet not functioning properly or experiencing connectivity issues. The final sample of 18 participants had 16 males and 2 females. 12 of the participants were in the 21-29 age group. Before starting the experiment, the participants reported their familiarity with the robot on a Likert scale ranging from 1-9, 1 being ``Not familiar at all" and 9 being ``Extremely familiar". Six participants reported 5 or less for familiarity with the robot, unlike the remaining 12, who were more familiar, reporting 6 or more. 

\section{RESULTS}
\label{sec:results}

This section presents the findings from our human-robot collaboration experiment, evaluating the impact of LED signals and emotional displays on human behaviour and perception. The results are organized to address the four hypotheses: \emph{H1} — emotional displays with LED signals increase collision anticipation; \emph{H2} —emotional displays with LED signals increase robot intent communication clarity; \emph{H3} — emotional displays with LED signals improve perceived robot competence; and \emph{H4} — emotional displays with LED signals alter human behavior compared to LED signals alone. 

To ensure the reliability of the questionnaire responses, we calculated McDonald's Omega ($\omega$) to assess internal consistency. $\omega$ values range from $0$ to $1$, where the values $\geq 0.9$ indicate excellent reliability, $0.8-0.9$ indicate good reliability, $0.7-0.8$ reflect acceptable reliability and the values $< 0.7$ suggest poor reliability. We conducted a Shapiro-Wilk normality test to assess whether the data followed a normal distribution. If the data followed a normal distribution, repeated measures ANOVA tests were used for comparisons, along with Bonferroni post-hoc tests for multiple comparisons when ANOVA was statistically significant. In cases where the data did not follow a normal distribution, Friedman's ANOVA tests were conducted, followed by Wilcoxon signed-rank tests to pinpoint the group(s) responsible for any significant findings.  For repeated measures ANOVA, we performed Mauchly's test of sphericity to assess whether the assumption of sphericity was met. If the test was statistically significant ($p<0.05$), indicating the assumption of sphericity had been violated, the Greenhouse-Geisser correction was applied.

\subsection{Questionnaire Data Analysis}
Questionnaire after each round (see Table~\ref{Tab:Questionnaire}) assessed participant perceptions of collision anticipation, communication clarity, and robot competence. McDonald’s Omega ($\omega$)) confirmed internal consistency: $\omega$ = $0.879$ (Condition 1), $0.935$ (Condition 2), and $0.954$ (Condition 3), indicating good to excellent reliability.

\subsubsection{Collision Anticipation}
\begin{figure}[h]
    \centering
    \includegraphics[width=0.7\linewidth]{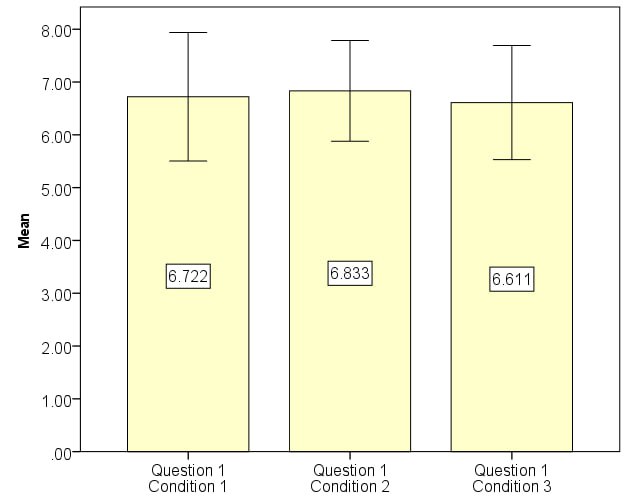}
    \caption{Mean scores and standard errors for anticipation of potential collision across the 3 conditions, showing no significant difference}
    \label{fig:Q1}
\end{figure}

Participants rated the statement “I was able to anticipate potential collisions with the robot” on a Likert-type scale with values ranging from 0 to 9. Shapiro-Wilk tests showed non-normal data ($p<0.05$). Friedman’s ANOVA revealed no significant differences across conditions ($p = 0.717$), with means of $6.722 \pm 2.445$ (Condition 1), $6.833 \pm 1.917$ (Condition 2), and $6.611 \pm 2.173$ (Condition 3) as shown in Figure \ref{fig:Q1}. Thus, \emph{H1} is not supported; that is, emotional displays did not significantly enhance collision anticipation beyond LED signals alone.

\subsubsection{Communication Clarity}
\begin{figure}[h]
    \centering
    \includegraphics[width=0.7\linewidth]{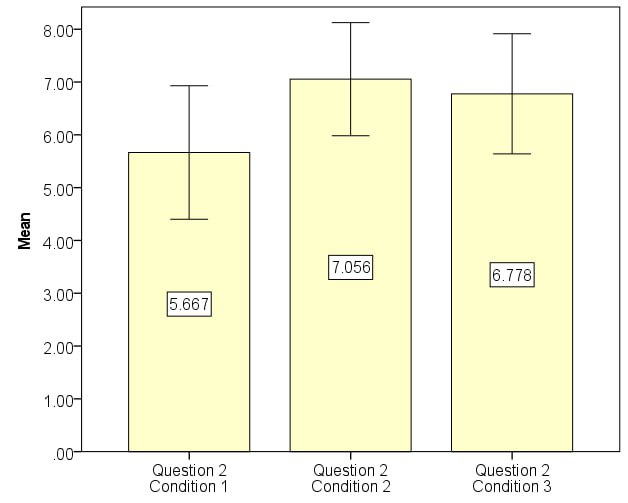}
    \caption{Mean scores and standard errors for robot communication intent across the three conditions, showing no significant difference}
    \label{fig:Q2}
\end{figure}

Participants rated the statement “The robot communicated its intent clearly” on a Likert-type scale with values ranging from 0 to 9. Shapiro-Wilk tests indicated non-normality for Conditions 2 and 3 ($p < 0.05$). Friedman's ANOVA did not show a significant effect ($p = 0.175$), with means of $5.667 \pm 2.544$ (Condition 1), $7.056 \pm 2.155$ (Condition 2), and $6.778 \pm 2.290$ (Condition 3) as shown in Figure~\ref{fig:Q2}. Thus, \emph{H2} is not supported; that is, emotional displays did not significantly improve the robot's intent communication clarity beyond LED signals alone.

\subsubsection{Perceived Competence}
\begin{figure}[h]
    \centering
    \includegraphics[width=\linewidth]{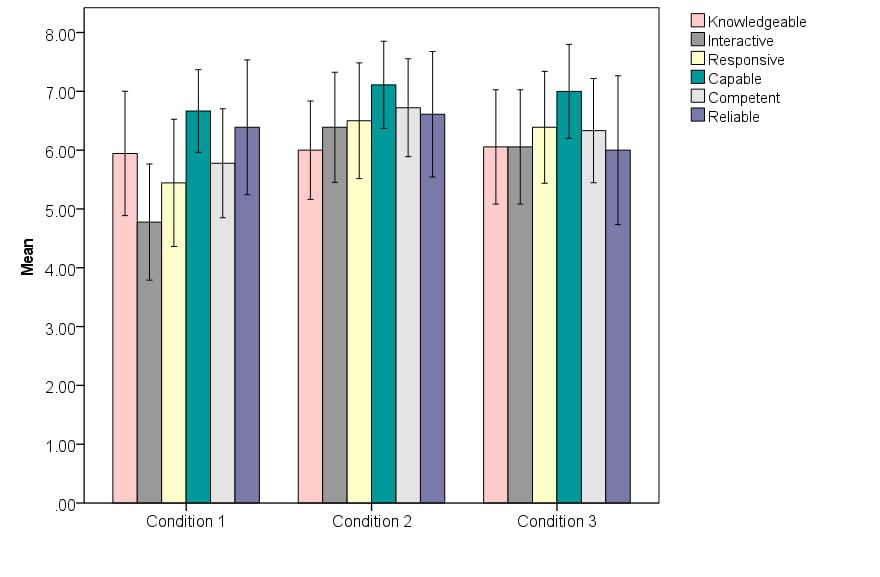}
    \caption{Means and standard errors for human-perceived competence of the robot showing that the robot was significantly more responsive in condition 2 than in condition 1}
    \label{fig:pc}
\end{figure}

We collected participants’ impressions of the robot in terms of six characteristics (Knowledgeable, Interactive, Responsive, Capable, Competent, and Reliable). Figure \ref{fig:pc} shows the means and standard errors of the responses. As shown in Table \ref{Tab:Prec}, Shapiro-Wilk tests showed mixed normality. For “Interactive”, Friedman's ANOVA was significant ($p = 0.019$), with Condition 2 ($6.389 \pm 1.883$) higher than Condition 1 ($4.778 \pm 1.987$), shown by Wilcoxon tests ($p < 0.05$). Other items (e.g., “Competent”: $5.778 \pm 1.865$, $6.722 \pm 1.674$, $6.333 \pm 1.782$) showed no significance ($p > 0.05$). Thus, \emph{H3} is partially supported, that is, emotional displays enhanced perceived interactivity but not overall competence beyond LED signals alone.

\begin{table}[h]
\centering
\caption{Testing the significance in participants’ answers across the three conditions}

\begin{tabular}{lcccc}
\hline
\textbf{Conditions}
 &
  \textbf{1} &
  \textbf{2} &
  \textbf{3} &
  \multirow{2}{*}{\textbf{\begin{tabular}[c]{@{}c@{}}P-value\\ ANOVA\end{tabular}}} \\ \cline{1-4}
 &
  \textbf{\begin{tabular}[c]{@{}c@{}}Mean\\ ± SD\end{tabular}} &
  \textbf{\begin{tabular}[c]{@{}c@{}}Mean\\ ± SD\end{tabular}} &
  \textbf{\begin{tabular}[c]{@{}c@{}}Mean\\ ± SD\end{tabular}} &
   \\ \hline
\textbf{Knowledgeable} &
  \begin{tabular}[c]{@{}c@{}}5.944\\ ± 2.127\end{tabular} &
  \begin{tabular}[c]{@{}c@{}}6.00\\ ± 1.680\end{tabular} &
  \begin{tabular}[c]{@{}c@{}}6.056\\ ± 1.955\end{tabular} &
  \begin{tabular}[c]{@{}c@{}}Repeated\\ Measures\\ 0.971 
  \end{tabular} \\
\textbf{Interactive} &
  \begin{tabular}[c]{@{}c@{}}4.778\\ ± 1.987{$^a$}\end{tabular} &
  \begin{tabular}[c]{@{}c@{}}6.389\\ ± 1.883{$^b$}\end{tabular} &
  \begin{tabular}[c]{@{}c@{}}6.056\\ ± 1.955{$^a$}{$^b$}\end{tabular} &
  \begin{tabular}[c]{@{}c@{}}Repeated\\ Measures\\ 0.019*\end{tabular} \\
\textbf{Responsive} &
  \begin{tabular}[c]{@{}c@{}}5.444\\ ± 2.175\end{tabular} &
  \begin{tabular}[c]{@{}c@{}}6.500\\ ± 1.978\end{tabular} &
  \begin{tabular}[c]{@{}c@{}}6.389\\ ± 1.914\end{tabular} &
  \begin{tabular}[c]{@{}c@{}}Friedman’s\\ 0.056\end{tabular} \\
\textbf{Capable} &
  \begin{tabular}[c]{@{}c@{}}6.667\\ ± 1.414\end{tabular} &
  \begin{tabular}[c]{@{}c@{}}7.111\\ ± 1.491\end{tabular} &
  \begin{tabular}[c]{@{}c@{}}7.00\\ ± 1.609\end{tabular} &
  \begin{tabular}[c]{@{}c@{}}Friedman’s\\ 0.390\end{tabular} \\
\textbf{Competent} &
  \begin{tabular}[c]{@{}c@{}}5.778\\ ± 1.865\end{tabular} &
  \begin{tabular}[c]{@{}c@{}}6.722\\ ± 1.674\end{tabular} &
  \begin{tabular}[c]{@{}c@{}}6.333\\ ± 1.782\end{tabular} &
  \begin{tabular}[c]{@{}c@{}}Friedman’s\\ 0.205\end{tabular} \\
\textbf{Reliable} &
  \begin{tabular}[c]{@{}c@{}}6.389\\ ± 2.304\end{tabular} &
  \begin{tabular}[c]{@{}c@{}}6.611\\ ± 2.146\end{tabular} &
  \begin{tabular}[c]{@{}c@{}}6.00\\ ± 2.544\end{tabular} &
  \begin{tabular}[c]{@{}c@{}}Friedman’s\\ 0.292\end{tabular} \\ \hline
\multicolumn{5}{l}{\begin{tabular}[c]{@{}l@{}}*Statistically significant $p < 0.05$.\\ {$^a$}{$^,$}{$^b$} : denote statistically significant differences between Condition 1 and \\ Condition 2 within the same row\end{tabular}} \\ \hline
\end{tabular}

\label{Tab:Prec}
\end{table}

\subsubsection{Comfort and Intent Perception}
\begin{figure*}[htbp]
    \centering
    \begin{subfigure}[b]{0.3\textwidth}
        \includegraphics[width=\textwidth]{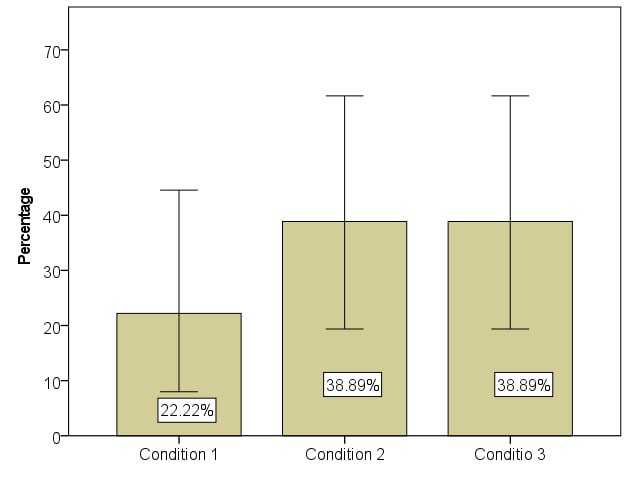}
        \caption{ Percentage of participants reporting clear robot intent across rounds}
        \label{fig:a}
    \end{subfigure}
    \hfill
    \begin{subfigure}[b]{0.3\textwidth}
        \includegraphics[width=\textwidth]{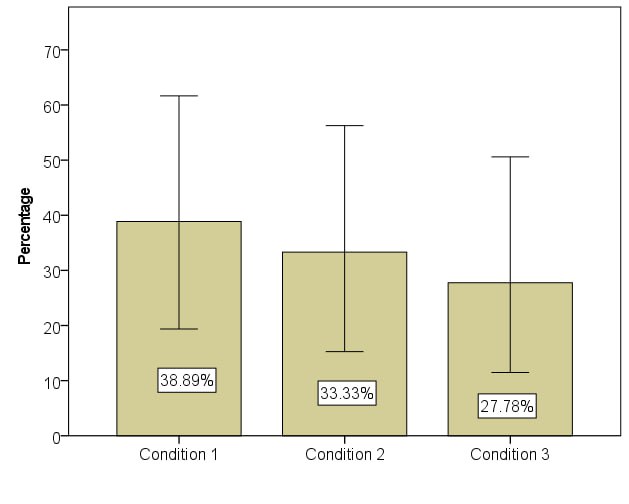}
        \caption{ Percentage of participants reporting most comfortable round}
        \label{fig:b}
    \end{subfigure}
    \hfill
    \begin{subfigure}[b]{0.3\textwidth}
        \includegraphics[width=\textwidth]{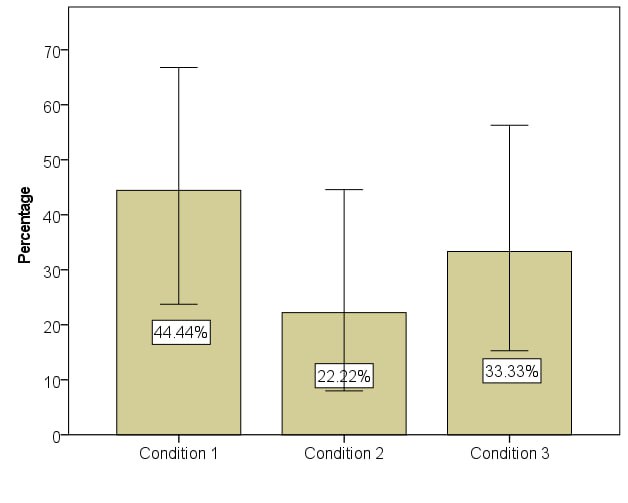}
        \caption{ Percentage of participants reporting the most uncomfortable round}
        \label{fig:c}
    \end{subfigure}
    \caption{Chi-square goodness of fit test to examine if there is significance in proportions of  participants’ perception of robot intent and comfort levels across conditions}
    \label{fig:final-questions}
\end{figure*}

We asked the participants three questions at the end of the experiment to identify the condition in which the robot’s intent was most clear, they felt the most comfortable, and they felt the most uncomfortable. The results are visualised in Figures~\ref{fig:final-questions}, adjusted for the different order of conditions such that each round number represents the corresponding signal condition. Chi-square tests on these post-experiment questions showed no significant differences ($p> 0.05$), suggesting consistent perceptions across conditions. 

\subsubsection{Open-ended Question}
The participants' responses to the open-ended question at the end of the experiment (Q4 -- Do you have anything else to say?) highlighted diverse participant experiences with the robot’s
behaviour and adaptability. Some participants found the robot’s actions predictable and
consistent, with comments such as, ``Consistent. The behaviour is predictable, and it’s easy
to work around it.'' Some participants reported being highly focused on the task, with one
noting, ``I was very concentrated on the task, but the face of the robot made me more cautious
about its movements.'' Others admitted to paying minimal attention to the robot, with one stating, ``I barely paid attention to it and stayed within a region where I knew I would not collide.'' Visual cues played a significant role in shaping these perceptions, with some participants pointing out challenges, such as, ``I noticed the missing colour on the tablet, which made it harder to understand what the robot would do." These mixed responses suggest variability in how participants interpreted and interacted with the system. The graphical user interface (GUI) and interactive elements received
positive feedback for improving user understanding and anticipation. Participants appreciated features like the robot’s waiting behaviour, with one stating, ``GUI helps me a lot
to understand the robot’s behaviour. When it was its turn, it waited for me as much as I
wanted, which is great." Similarly, the animations were praised for their effectiveness, as highlighted in the comment, ``The interactive display with animations was super helpful
and made a big difference in my anticipation of the robot." Overall, feedback underscores the importance of clear and consistent visual communication in human-robot interaction, and how it should be explored further, as well as the need for further testing to enhance perceived adaptability and user experience.

\subsection{Position Tracking Analysis}
Position tracking data was collected per participant in three conditions. The variables included the time spent in each round, 3d positions of the robot’s end effector and participant’s hand, the distance between them, and the LED colour (green or red).

\subsubsection{Collision Avoidance Behavior}
We calculated the time spent by the human hand in the robot's movement path in each condition, for the yielding and unyielding modes. If the participants were prone to avoiding potential collisions in conditions with emotional displays, the time spent by them in the robot's movement path would be lower in Conditions 2 and 3 as compared to Condition 1. For the unyielding mode (Red LED), the means and standard deviations were: $506.833 \pm 289.458$ms (Condition 1), $682.944 \pm 223.332$ms (Condition 2), and $580.389 \pm 245.268$ms (Condition 3). Shapiro-Wilk tests indicated non-normality ($p < 0.05$), so Friedman’s ANOVA was applied, showing a non-significant result ($p = 0.128$). For the yielding mode, the means and standard deviations were: $1007.50 \pm 448.094$ms (Condition 1), $879.278 \pm 528.620$ms (Condition 2), and $933.944 \pm 406.844$ms (Condition 3), also non-normal ($p < 0.05$), with Friedman’s ANOVA not showing any significance ($p = 0.179$). 

We calculated the average distance between the participant’s hand and the robot’s end effector in each condition. If the participants were prone to avoiding potential collisions in conditions with emotional displays, the average distance between their hand and the robot's end effector would be higher in conditions 2 and 3 as compared to Condition 1.  The means and standard deviations of the average distance were: $0.576 \pm 0.041$m (Condition 1),  $0.560 \pm 0.030$m (Condition 2), and $0.569 \pm 0.048$m (Condition 3). Shapiro-Wilk tests confirmed normality (Condition 1: $W = 0.916$, $p = 0.110$; Condition 2: $W = 0.988$, $p = 0.996$; Condition 3: $W = 0.969$, $p = 0.783$), and Mauchly’s test showed no sphericity violation ($p = 0.704$). Repeated Measures ANOVA did not indicate a significant effect ($p = 0.461$). 
Thus, there was no significant difference between the conditions in terms of the participants' collision avoidance behaviour.

\subsubsection{Average Completion Duration}
We calculated the duration of task completion for each condition to test the effect of emotional displays on the efficiency of the participants. The means and standard deviations of the task execution time were: $110.8 \pm 14.3$s (Condition 1), $114.7 \pm 16.0$s (Condition 2), and $113.1 \pm 15.2$s (Condition 3). Shapiro-Wilk tests confirmed normality (Condition 1: $W = 0.989$, $p = 0.998$; Condition 2: $W = 0.977$, $p = 0.920$; Condition 3: $W = 0.978$, $p = 0.926$), and Mauchly’s test indicated no sphericity violation ($p = 0.518$). Repeated Measures ANOVA did not indicate a significant effect ($p = 0.175$). These results show that emotional display did not affect the participants' efficiency in terms of the task completion duration.

\section{DISCUSSION}
We investigated the impact of nonverbal communication techniques on human-robot collaboration, focusing on LED signals and emotional displays. The results suggest that, while emotional displays enhance the perceived interactivity of the robot, they do not significantly improve collision anticipation or communication clarity beyond what LED signals provide. Furthermore, while participants responded positively to emotional displays in their open-ended responses, their overall task performance and perception of safety did not change significantly. These findings indicate that, while emotional displays can be beneficial for engagement, their role in practical shared workspaces remains limited without further refinement, additional sensory feedback, or multimodal cues.

This study was limited to 18 participants, primarily from an academic setting. In addition, the experiment was conducted in a confined space. Future experiments could expand to a larger set of participants: industrial workers, engineers, various age groups, and individuals with varying levels of robotics experience and a larger space with a mobile manipulator, to better assess the generalizability of the findings.

Furthermore, to improve and refine nonverbal communication in human-robot collaboration, several research directions could be explored. Future work could investigate multimodal communication by integrating gaze cues along with LED signals and facial expressions. Gaze-based intent signalling, for example, has been shown to improve predictability in HRI~\cite{wang_gaze_2019} and could complement the existing emotional display system. The current system uses predefined reactions to human movement, but a real-time adaptive system could significantly enhance responsiveness. Using machine learning or reinforcement learning, the robot could dynamically adjust its communication strategy based on the behaviour of the participant, providing more personalised and effective feedback. In addition, human adaptation to robot behaviour may evolve. A longitudinal study examining how participants respond to nonverbal cues over multiple sessions or weeks would provide deeper insights into whether emotional displays improve efficiency and trust in the long run.

\addtolength{\textheight}{-8cm}   

\section*{ACKNOWLEDGMENT}
We thank all the people who participated in the experiment. We also thank every individual who provided technical support and assistance throughout this project.

\bibliographystyle{IEEEtran}
\bibliography{name}

\end{document}